\documentclass{amia}

\usepackage{lipsum}
\usepackage[utf8]{inputenc} 
\usepackage[T1]{fontenc}    
\usepackage{hyperref}       
\usepackage{url}            
\usepackage{booktabs}       
\usepackage{amsfonts}       
\usepackage{nicefrac}       
\usepackage{microtype}      
\usepackage{xcolor}         
\usepackage{enumitem}
\usepackage{amsmath}
\usepackage{mathrsfs}
\usepackage{subfigure}
\usepackage{graphicx}
\usepackage{amsthm}
\usepackage{color}
\usepackage{mathrsfs}
\usepackage{amsfonts}
\usepackage{algorithm, algpseudocode}
\usepackage{wrapfig}
\usepackage{blindtext}
\usepackage{caption}
\usepackage{makecell}
\usepackage{fontawesome}
\usepackage{tikz}

\def\Algnameunderline{\underline{S}oft \underline{P}rompt-Bas\underline{e}d \underline{C}alibration}
\def\Algnameabbr{SPeC}
\newcommand{\nerbox}[1]{\tikz[baseline=(X.base)]\node [draw=black!50,fill=red!15,rectangle,inner sep=1pt, rounded corners=3pt](X) {#1};}

\begin{document}

\title{\Algnameabbr{}: A Soft Prompt-Based Calibration on Performance Variability of Large Language Model in Clinical Notes Summarization}

\author{Yu-Neng Chuang$^1$, Ruixiang Tang$^1$, Xiaoqian Jiang$^2$, PhD, and Xia Hu$^1$, PhD }
\institutes{
    $^1$ Rice University, Houston, TX; $^2$ University of Texas Health Science Center, Houston, TX
}

\maketitle

\section*{Abstract} \textit{Electronic health records (EHRs) store an extensive array of patient information, encompassing medical histories, diagnoses, treatments, and test outcomes. These records are crucial for enabling healthcare providers to make well-informed decisions regarding patient care. Summarizing clinical notes further assists healthcare professionals in pinpointing potential health risks and making better-informed decisions. This process contributes to reducing errors and enhancing patient outcomes by ensuring providers have access to the most pertinent and current patient data. Recent research has shown that incorporating instruction prompts with large language models (LLMs) substantially boosts the efficacy of summarization tasks. However, we show that this approach also leads to increased output variance, resulting in significantly distinct summaries even when instruction prompts share similar meanings. To tackle this challenge, we introduce a model-agnostic \Algnameunderline{} (\Algnameabbr{}) pipeline that employs soft prompts to lower variance while preserving the advantages of prompt-based summarization. Experimental findings on multiple clinical note tasks and LLMs indicate that our method not only bolsters performance but also effectively regulates variance across different LLMs, providing a more consistent and reliable approach to summarizing critical medical information.}

\section{Introduction}
Electronic health records (EHRs) have brought about a revolutionary change in the accessibility and utilization of patient information by healthcare providers. This transformation has impacted the decision-making process of medical professionals, enabling them to make informed decisions about patient care~\cite{wagner2020augmented}. Among the many benefits of EHRs, the ability to summarize clinical notes is of particular importance \cite{pivovarov2015automated}, as it enables healthcare providers to quickly identify potential health risks and make better-informed decisions. By presenting the most relevant and up-to-date patient information, clinical note summarization ultimately contributes to a reduction in diagnosis errors and an improvement in patient outcomes \cite{wang2021systematic}. However, manual summarization of clinical findings and reports into summaries is both time-consuming and prone to errors~\cite{gershanik2011critical}. Moreover, given the volume and complexity of the data, even experienced clinicians can inadvertently overlook significant aspects of the patient's condition. Thus, there is a pressing need to develop automated methods for generating summaries, enhancing both efficiency and accuracy in patient care.

Recent developments in the field of natural language processing, particularly the advent of large language models (LLMs), have showcased a substantial potential for enhancing the efficiency of automated clinical note summarization \cite{cai2021chestxraybert, xiao2018opportunities, choi2018mime}. It's been noted that these models show remarkable ability for aligning input instruction \cite{wei2022emergent}, thereby suggesting the integration of text prompts as a viable approach for improving LLM performance in summarization tasks \cite{liu2023pre}. For instance, a carefully crafted instruction prompt such as 'Summarize the key findings in the patient's medical history' effectively directs the LLM's focus toward extracting the most pertinent information from the clinical notes. However, designing an effective text prompt—termed 'prompt engineering'—is challenging. It requires precision, informativeness, and a confluence of knowledge from several domains~\cite{zhang2022neural}. Moreover, our study reveals that the use of manually designed instruction prompts can induce increased output variance. Even minor modifications to the prompts can result in significant variations in the summarization outcomes, as demonstrated in Table \ref{tab:chatgpt-exmaple}. This instability in the performance of LLM-generated prompts may limit the ability of non-NLP experts to effectively leverage LLMs for their intended tasks.

In response to this challenge, we present a model-agnostic approach named \Algnameunderline{} (\Algnameabbr{}), designed to address the issue of performance variability in clinical notes summarization \cite{raffel2020exploring}. Unlike conventional discrete text prompts, where each input token carries a definite meaning, soft prompts are flexible and learnable tokens devoid of pre-defined significance \cite{lester2021power}. This flexibility enables the model to learn specific parameters tailored for them. By leveraging soft prompts, our approach aims to mitigate variance while retaining the benefits of discrete prompt-based summarization. Specifically, we propose a soft prompt encoder to interact soft prompts with discrete text prompts throughout the token embedding space. Our proposed soft prompt encoder is a zero-shot learning model that does not require any of the golden summarization references as ground truth labels during the training phase. Our experimental findings demonstrate that \Algnameabbr{} not only improves overall performance but also effectively reduces variance across different LLMs. For instance, \Algnameabbr{} deduces up to 43.1\% of the performance variability in the Flan-T5 Model~\cite{chung2022scaling}. This results in a more uniform and reliable solution for summarizing crucial medical information, ultimately supporting healthcare professionals in making well-informed decisions and providing optimal patient care. The success of \Algnameabbr{} in addressing the issue of performance variability has significant implications for the future of clinical note summarization and the broader application of LLMs in the healthcare domain.\\

\begin{table}[t!]
    \centering
    \small
    \begin{tabular}{c | p{13cm}}
        \toprule
        \multicolumn{2}{c}{\textbf{Radiology Report Data}} \\
        \midrule
        Original Report & There has been interval removal of a left-sided PICC line. Cardio mediastinal and hilar contours are unchanged. The Pleurx catheter is seen extending posteriorly. There is a loculated pleural air inclusion on the right. There is no pneumothorax. Sternotomy wires are aligned. \\
        \midrule
        Doctors' Finding & Loculated pleural air inclusion on the right. Pleurx catheter in good position.\\
        \midrule\midrule
        \multicolumn{2}{c}{\textbf{\faThumbsOUp ~ Clinical Summarization by LLM-generated Prompt ONE ~\faThumbsOUp}} \\
        \midrule
        Prompt & Summarize the patient's current medical status, medical history, test results, and treatment plan from the clinical note. \\
        \midrule
        Summarization & Persistent right-sided loculated pleural air with no pneumothorax\\
        \midrule\midrule
        \multicolumn{2}{c}{\textbf{\faThumbsODown ~ Clinical Summarization by LLM-generated Prompt TWO ~\faThumbsODown}} \\
        \midrule
        Prompt & Based on the patient's clinical notes, can you give me a concise overview of their diagnosis, treatment plan, and prognosis? \\
        \midrule
        Summarization & Right pleural opacity concerning for pneumonia. Pleurx catheter has been removed \\
        \bottomrule
    \end{tabular}
    \caption{An example of performance variability of LLM-generated prompt under Flan-T5 model.}
    \label{tab:chatgpt-exmaple}
\end{table}

\section{Preliminaries}

\subsection{Clinical Note Summarization}
Clinical note summarization focuses on producing a succinct overview of a patient's medical history, encompassing aspects such as their medical background, symptoms, diagnoses, treatments, and outcomes. Current methods can be divided into extractive \cite{gupta2010survey} and abstractive techniques \cite{lin2019abstractive}. Extractive approaches pinpoint and assemble relevant sentences or phrases from the original clinical documentation to create a summary. Conversely, recent research has primarily focused on abstractive techniques, which generate summaries by rephrasing and fusing information from the source text, utilizing large language models \cite{see2017get, cai2021chestxraybert, lewis2020bart}. Challenges associated with clinical note summarization consist of navigating the intricate and variable nature of medical terminology, addressing noise and inaccuracies in the input text, and grasping the context and subtleties of clinical information. Furthermore, clinical notes often include unstructured data, such as free text or handwritten entries, which pose difficulties in processing and summarization \cite{jain2022survey}. 

\subsection{Benefits from Prompt Engineering on LLMs}
Prompt engineering is a concept in natural language generation, which involves creating prompts that steer LLMs toward desirable or useful results. A primary advantage of employing prompt engineering lies in its capacity to enhance the caliber of the outputs produced by LLMs. By crafting prompts that direct the model to concentrate on pertinent aspects of the input, researchers can generate more precise and consistent outputs. 
In general, there are two different kinds of prompts: the discrete textual prompt and the soft prompt~\cite{liu2023pre}. The first one aims to generate the prompts corresponding to a natural language phrase, such as providing a sequence of words. The second one performs prompting directly in the embedding space of the LLMs instead of involving any human consumption, such as providing the trainable vectors. With the discrete textual prompt, recent study~\cite{Zhou2022-td, shin2020autoprompt, li2021prefix}, utilizing automatic prompt engineering (APE) can boost the accuracy and comprehensiveness of LLMs, as well as augment the effectiveness of few-shot learning. Some research~\cite{wang2022promptehr, vu2021spot, kim2023prompt} also elaborates that soft prompt tuning can effectively improve the performance in electronic healthcare records generation and other medical applications~\cite{zhang2023pheme, yuan2023llm}. However, these two prompt designing methods are still too challenging for non-NLP experts to exploit directly. Notably, a haphazard prompt design may lead to even worse performance of LLMs, which causes the final results of LLMs to be unstable and uncontrollable. This fluctuating situation eventually set a barrier for prompt engineering to be widely applied to healthcare applications. 

\begin{table}[t!]
    \centering
    \small
    \begin{tabular}{l}
        \toprule
        \multicolumn{1}{c}{\textbf{Selected LLM-generated Prompts}} \\
        \toprule
        (1) Please summarize the patient's medical history and current symptoms. \\
        (2) Please summarize the key findings and treatment plan from the patient's recent clinical notes. \\
        (3) Can you please provide a brief summary of the patient's current condition and medical history based on the clinical notes? \\
        (4) Can you summarize the patient's diagnosis, symptoms, and any relevant test results from their most recent visit? \\
        (5) Can you provide a brief summary of the patient's medical history and current condition based on their clinical notes? \\
        (6) Based on the patient's clinical notes, can you give me a concise overview of their diagnosis, treatment plan, and prognosis? \\
        (7) Summarize the following clinical notes. \\
        (8) How would you summarize the key information and findings documented in the clinical note? \\
        (9) Please summarize the patient's medical history, including any chronic conditions, medications, and allergies. \\
        (10) Summarize the patient's current medical status, medical history, test results, and treatment plan from the clinical note. \\
        \bottomrule
    \end{tabular}
    \caption{Examples of selected LLM-generated prompts for clinical notes summarization task.}
    \label{tab:chatgptprompt}
\end{table}

\subsection{Variability on LLM-generated Prompts}
While LLMs have shown great promise in generating natural language outputs, prompting these models can be a delicate process.
Even slight modifications to the prompts can cause significant variations in model predictions, making prompting a brittle process~\cite{Arora2022-dj}.
To overcome this challenge, researchers have proposed various methods for prompt engineering. For instance, PromptChainer is a tool that allows implementing, improving, and testing individual nodes of LLM prompts and also supports running the chain end-to-end~\cite{Wu2022-dw}. Additionally, Mishra et al. have suggested reframing prompts to make them more amenable to LLM language and generate coherent  answers\cite{Mishra2021-bl}. There is still a lack of effective general-purpose prompt engineering techniques to mitigate instability issues in AI language models.

\section{Data and Problem Description}

\noindent
\textbf{Dataset.} MIMIC-CXR~\cite{johnson2019mimic} is a large dataset of chest radiographs (X-rays) and their corresponding radiology reports designed to support research in medical image analysis and natural language processing. It contains over 377,000 images and their associated reports, which were collected from patients admitted to the intensive care units of the Beth Israel Deaconess Medical Center in Boston between 2001 and 2012. The MIMIC-CXR dataset has been widely used for tasks such as image classification, disease detection, and report generation in medical research. A typical radiology report generally comprises three sections: (1) a background section with the patient's and exam's general information, (2) a findings section that presents the examination's details, and (3) an impressions section that summarizes the findings~\cite{kahn2009toward}. In this work, we focus on the clinical notes summarization task. The ``findings" are treated as the input clinical notes, and the ``impressions" are the ground-truth summarization of radiology notes. To obtain access to the MIMIC-CXR dataset, we complied with the necessary requirements by completing the CITI Data training courses, as mandated by the data providers. Subsequently, we obtained the dataset by downloading it from PhysioNet, a platform that hosts various biomedical datasets for research purposes. 

\noindent
\textbf{Prompts Collection from LLMs.} With impressive results in natural language understanding and generation, LLMs are able to assist experts that are not familiar with language models' architecture in creating the appropriate prompts for fitting their tasks on specific domains. For example, we can use ChatGPT to automatically generate the prompts as the input for other pre-train LLMs. As LLMs' performance largely depends on prompt quality but without finetuning, an automatic prompt generation can assist non-NLP experts in making better use of the powerful pre-train LLMs. In this work, we exploit ChatGPT as an example to demonstrate the prompt-generating process, which tries to simulate the use case of non-NLP experts. ChatGPT is a large language model developed by OpenAI, based on the GPT-3.5 architecture~\cite{gpt4}. Specifically, ChatGPT is capable of generating human-like responses to a wide variety of prompts, including questions and writing prompts. With impressive results in paragraph generation, ChatGPT is able to assist experts that are not familiar with GPT architecture. In this work, ten ChatGPT-generated prompts are randomly collected for the clinical notes summarization task, as shown in Table~\ref{tab:chatgptprompt}. To simulate the non-NLP experts on prompt design, ``What is a good prompt for Clinical Notes Summarization?" is adopted as the simple input question and collect the return answers from ChatGPT. All generating processes are ensured to complete in new dialogues of ChatGPT, ensuring the independence of each prompt-generating process. 

\noindent
\textbf{Problem Objectives} The objective is to develop a calibrated model that can reduce the performance variability in clinical note summarization while being robust to the quality impacts of LLM-generated prompts. The use of discrete text prompts can potentially bolster the adaptability of LLMs for a variety of downstream tasks. However, prompt-based adaptation is not without its limitations. The task description is error-prone and highly reliant on human input, as outlined by Lester et al. (2021)~\cite{lester2021power}. Lately, the trend is shifting towards utilizing LLM-generated prompts to alleviate the laborious process of crafting discrete prompt texts. Nevertheless, the quality of these LLM-generated prompts can be highly inconsistent, thereby substantially influencing the performance of pre-trained LLMs in clinical note summarization. Therefore, an ideal calibration model should effectively mitigate this variance while maintaining the performance advantages of LLM-generated prompts.

\begin{figure}[t]
\centering
    \includegraphics[width=0.9\textwidth]{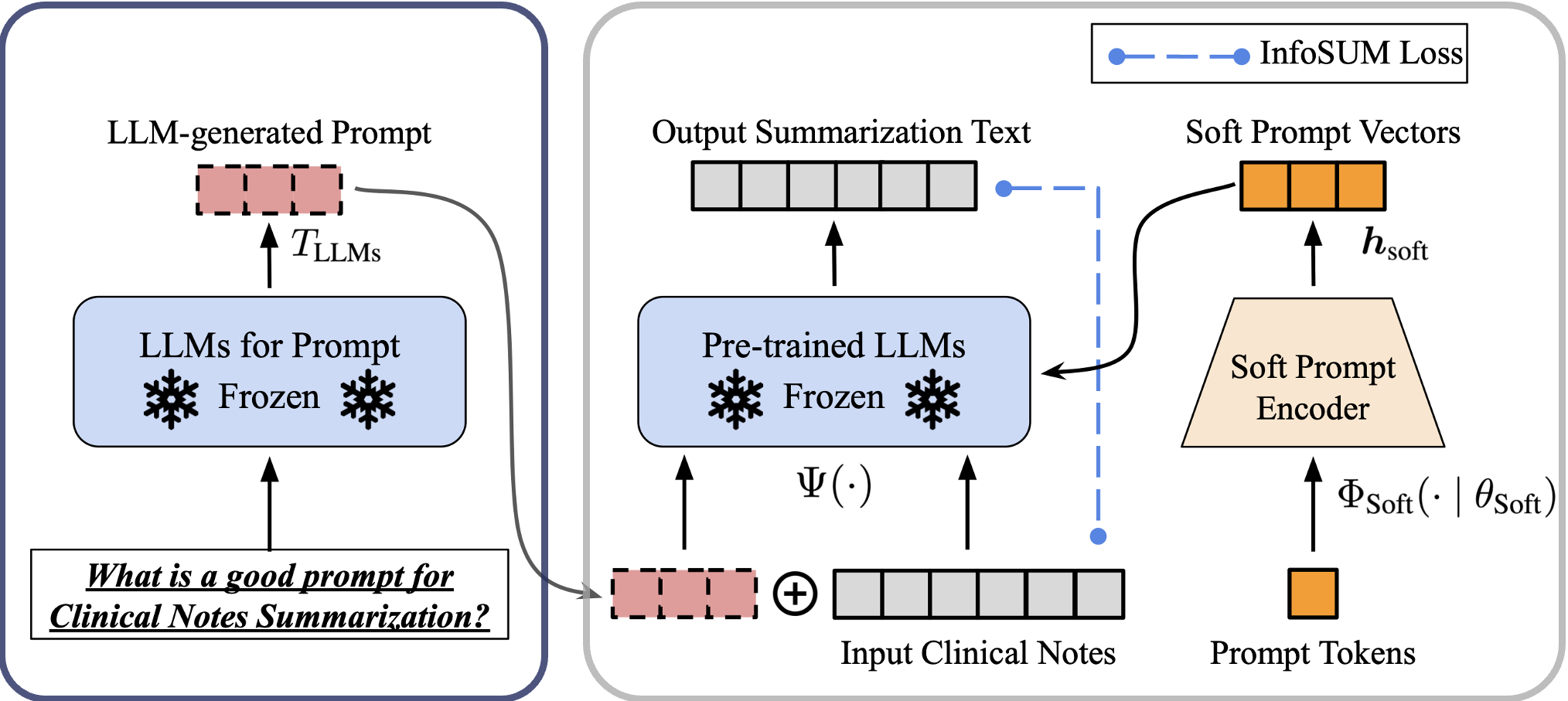} 
    \caption{An overview of the pipeline for \Algnameabbr{} framework with LLM-generated prompts. "LLMs for Prompt" (navy blue part) is in charge of automatically generating prompts for clinical note summarization. The requirement for LLMs selection in the first part is to ensure that chosen LLMs obtain superior text generation capability, such as GPT4 and LLama are good candidates. "Pre-train LLMs" (gray part) is responsible for yielding the summarization of a given clinical note accompanied by the LLM-generated prompts from the blue part. "Pre-train LLMs" are required to perform well in text summarization tasks.}
\label{fig:framework} 
\end{figure}

\section{Methodology}

We systematically introduce the \Algnameunderline{} (\Algnameabbr{}) framework in this section. Figure~\ref{fig:framework} illustrates the overall pipeline of our framework. In particular, we first collect the LLM-generated prompts from LLMs. Then the purposed soft prompt encoder is further trained with \textsc{InfoSUM} loss to mitigate the uncertain performance brought by LLM-generated prompts. Specifically, \Algnameabbr{}  is a zero-shot and model-agnostic framework with a soft prompt encoder, incorporating different transformer-based language models.

\subsection{Soft Prompt Encoder} 
The objective of the soft prompt encoder is to reduce performance variability caused by LLM-generated prompts by utilizing soft prompts without any retraining or finetuning. 
This approach aims to enhance the robustness of LLM-generated prompts without significant alterations to their length and contents, as the length of LLM-generated prompts is restricted to be too long due to the LLMs' input constraints. 
Considering the quality of prompts generated by LLMs is often unpredictable and inconsistent, this can significantly impact the summarizing performance of clinical notes from pre-trained LLMs. To address this challenge, we introduce a soft prompt encoder that reduces the impact of the unpredictable quality of prompts generated by LLM. Motivated by prompt tuning concepts~\cite{lester2021power}, we want 
The soft prompt encoder modifies the soft prompt vectors to effectively reduce variability while incorporating the LLM-generated prompts to improve performance. In this work, we randomly select several words as the input soft prompt token from the existing token dictionary of pre-trained LLMs. The input soft tokens can be randomly selected as either a string or a sentence. For example, the soft prompt token can be set as \textit{``radiologist describe stable normality and abnormality exam"}. The corresponding soft prompt vectors are then developed by the soft prompt encoder under the given soft prompt token.
Formally, given a soft prompt token $\langle\text{Soft-tkn}\rangle$ and a soft prompt encoder $\Phi_{\text{Soft}}({\cdot} ~|~ \theta_\text{Soft}): \mathbb{R}^m \rightarrow \mathbb{R}^d$, the soft prompt vector $\boldsymbol{h}_{\text{soft}} \in \mathbb{R}^d$ can be denoted as:
\begin{equation}
    \boldsymbol{h}_{\text{soft}} = \Phi_{\text{Soft}}\big(\langle\text{Soft-tkn}\rangle ~|~ \theta_\text{Soft} \big)
    \label{eq:softenc}
\end{equation}
where $\Phi({\cdot} ~|~ \theta_\text{Soft})$ can be an encoder of arbitrary pre-trained transformer-based LLMs. During the inference phase, the soft prompt vector can be decoded with special invariant tokens with the frozen decoder of selected pre-trained LLMs. The generated invariant soft tokens can then be directly employed to incorporate LLM-generated prompts, thus ensuring low variability of the performance.

\begin{algorithm}[t!]
    \caption{Soft prompt encoder with \Algnameabbr{}}
    \begin{algorithmic}[1]
        \State {\bfseries Input:} 
      
            Frozen pre-trained encoder $\textit{Enc}(\cdot ~|~ \theta^{*}_\text{Enc})$ and frozen pre-trained decoder $\textit{Dec}(\cdot ~|~ \theta^{*}_\text{Dec})$,
    
            Trainable Base Soft prompt encoder $\Phi_{\text{Soft}}({\cdot} ~|~ \theta_\text{Soft})$,
            
            Original clinical notes input $T_\text{org}$,
            
            Collected Numerous ChatGPT-generated prompts $T_\text{chatgpt}$,
            
            Selected soft prompt token $\langle\text{Soft-tkn}\rangle$.
        
        \State {\bfseries Output:} 
        
            Soft prompt encoder $\Phi_{\text{Soft}}({\cdot} ~|~ \theta_\text{Soft})$.
        
        \While{not convergence}
          \State Generate soft prompt vectors $\boldsymbol{h}_{\text{soft}}$ from Equation~\ref{eq:softenc}.
          \State Update $\Phi_{\text{Soft}}({\cdot} ~|~ \theta_\text{Soft})$ with $T_\text{org}$ and $T_\text{chatgpt}$ to minimize \textsc{InfoSUM loss} given by Equation~\ref{eq:infoloss}. 
        \EndWhile
    \end{algorithmic}
    \label{alg:bmixup}
\end{algorithm}

\subsection{\textsc{InfoSUM Loss}}
In order to train the soft prompt encoder $\Phi_{\text{Soft}}({\cdot})$, we propose InfoSUM loss to ensure the generated soft prompt vector $\boldsymbol{h}_{\text{soft}}$ are able to degrade the variability of prompted input (i.e., input clinical notes with LLM-generated prompt). The \textsc{InfoSUM loss} aims to measure the distance between the embedding distribution of input clinical notes and the embedding distribution of output summarization under the frozen transformer-based LLMs $\Psi(\cdot)$. The transformer-based LLMs are typically composed of encoders $\textit{Enc}(\cdot ~|~ \theta_\text{Enc})$ and decoders $\textit{Dec}(\cdot ~|~ \theta_\text{Dec})$, where we denote as $\Psi(\cdot) = \textit{Dec}( \textit{Enc}(\cdot ~|~ \theta_\text{Enc}) ~|~ \theta_\text{Dec} )$.
Specifically, the primary objective of \textsc{InfoSUM Loss} is to minimize the divergence between the prompted input and output distributions underlying the effect of a soft prompt vector. Let $T_\text{org}$ be original input clinical notes and $T_\text{LLMs}$ be prompt. We formally define the embedding distribution of original clinical notes $p_{\Psi}^{\text{Org}}(\cdot)$ and summarization output $p_{\Psi}^{\text{Prmt}}(\cdot)$ as follows:
\begin{align}
    \begin{cases}
        \notag \text{Original Clinical Notes Emb} := p_{\Psi}^{\text{Org}}(~T_\text{org}~) = \textit{Enc}(T_\text{org} ~|~ \theta^{*}_\text{Enc}) \\
        \notag \text{Summarization Output Emb} := p_{\Psi}^{\text{Prmt}} \big( T_\text{org}, T_\text{LLMs}, \langle\text{Soft-tkn}\rangle \big) = \textit{Enc} \big(T_\text{org}, T_\text{LLMs} ~|~ \theta^{*}_\text{Enc} \big) \oplus \Phi_{\text{Soft}}\big(\langle\text{Soft-tkn}\rangle ~|~ \theta_\text{Soft} \big)
    \end{cases}
\end{align}
where $\theta^{*}_\text{Enc}$ denotes the frozen encoder parameters and $\oplus$ represents the mean value operation of two given embeddings.
Following the criteria, we can minimize the \textsc{InfoSUM loss} by purely updating the parameters of soft prompt encoder $\theta_\text{Soft}$ as follows:
\begin{align}
    \small
    \notag \mathcal{L}_\textsc{InfoSUM}\big(~ p_{\Psi}^{\text{Org}}(\cdot), ~p_{\Psi}^{\text{Prmt}}(\cdot) ~\big) =& \mathbf{I} \big[~ \textit{Enc}(p^{\text{Org}}_{\Psi}(~T_\text{org}~), ~p^{\text{Prmt}}_{\Psi}(T_\text{org}, T_\text{LLMs}, \langle\text{Soft-tkn}\rangle ) ~\big] \\
    =& \mathbf{I} \big[~ \textit{Enc}(T_\text{org} ~|~ \theta^{*}_\text{Enc}), ~\textit{Enc}(T_\text{org}, T_\text{LLMs} ~|~ \theta^{*}_\text{Enc}) \oplus \Phi_{\text{Soft}}\big(\langle\text{Soft-tkn}\rangle ~|~ \theta_\text{Soft} \big) ~\big]
    \label{eq:infoloss}
\end{align}
where $\mathbf{I}[\cdot]$ represents any distance measurement functions. In practice, the selected distance functions should follow similar measuring concepts of pre-trained LLMs, which can simultaneously ensure the summarization performance and learn a soft prompt vector properly for variability reduction. In this work, the soft prompt tokens are given as \textit{``radiologist describe stable normality and abnormality exam"}.

\subsection{Algorithm of \Algnameabbr{}}
The outline of \Algnameabbr{} is given in Algorithm~\ref{alg:bmixup}. \Algnameabbr{} learns a soft prompt vector $\boldsymbol{h}_{\text{soft}}$ according to the soft prompt encoder $\Phi_{\text{Soft}}({\cdot} ~|~ \theta_\text{Soft})$ in Equation~\ref{eq:softenc} (line 4), and optimize the soft prompt encoder with \textsc{InfoSUM Loss} in Equation~\ref{eq:infoloss} (line 5). The training iteration terminates when $\Phi_{\text{Soft}}({\cdot} ~|~ \theta_\text{Soft})$ is converged. In this work, the initial weights of base $\Phi_{\text{Soft}}({\cdot} ~|~ \theta_\text{Soft})$ is the same as frozen pre-trained encoder $\textit{Enc}(\cdot ~|~ \theta^{*}_\text{Enc})$, where \Algnameabbr{} only updates $\Phi_{\text{Soft}}({\cdot} ~|~ \theta_\text{Soft})$. After the training phase of  $\Phi_{\text{Soft}}({\cdot} ~|~ \theta_\text{Soft})$, the final clinical notes summarization $T_\text{summ}$ can be derived by the frozen pre-trained LLMs $\Psi(\cdot | \theta^{*}_\text{Enc}, \theta^{*}_\text{Dec})$ with a decoded soft prompt text $\textit{Dec}( \boldsymbol{h}_{\text{soft}} ~|~ \theta^{*}_\text{Dec})$ and prompted clinical notes $T_\text{org} + T_\text{LLM}$.

\section{Experiments}

In this section, we conduct experiments to analyze the performance of \Algnameabbr, which aims to answer the following three research questions.
\begin{itemize}[leftmargin=0.5cm]
    \item \textbf{RQ1:} How does \Algnameabbr{} perform on clinical notes summarization task in terms of its efficacy and variability? 
    \item \textbf{RQ2:} How effective does \Algnameabbr{} incorporate multiple transformer-based LLMs?
    \item \textbf{RQ3:} Does the selection of soft prompt token affects the efficacy and variability of \Algnameabbr{}?
\end{itemize}

\subsection{Experimental Settings}
In this section, we introduce the experimental settings and metrics for evaluating \Algnameabbr{} on the clinical notes summarization task. The implementation details and baseline settings are shown as follows.
    
\noindent
\textbf{Baselines.} To better demonstrate the efficacy in mitigating the variability of ChatGPT-generated prompts, we select three different transformer-based LLMs as the baselines, which are Flan-T5~\cite{chung2022scaling}, BART~\cite{lewis2020bart}, and Pegasus-xsum~\cite{zhang2019pegasus}. These models are based on the transformer architecture and follow an encoder-decoder structure, and these three models are popular and specifically designed for text summarization tasks.
Flan-T5 is trained using an instruction-alignment dataset that enables it to excel across diverse NLP tasks. BART is a denoising autoencoder for pretraining sequence-to-sequence models. Pegasus-xsum stands out as a pretraining model explicitly tailored for abstractive summarization tasks. All baseline models selected in their base weight version are finetuned with the MIMIC-CXR dataset, where default settings of the original papers decide the hyper-parameters on model training.

\noindent
\textbf{Evaluation Protocol.} For a fair comparison, all methods are tested under the same experimental settings as outlined below: the training set and testing set are randomly split from the original MIMIC-CXR dataset, where the training set obtains 91,544 clinical notes, and the testing set has 200 clinical notes. As for the evaluation metrics, three commonly used ROUGE metrics are chosen to verify the performance of \Algnameabbr{} with ground-truth reference $T_\text{ref}$ provided by the MIMIC-CXR dataset. Specifically, ROUGE (Recall-Oriented Understudy for Gisting Evaluation)~\cite{lin2004rouge} calculates the count of overlapping 'n-grams' between the model-generated summary $T_\text{sum}$ and the reference $T_\text{ref}$. An n-gram refers to a set of tokens/words, where a unigram (1-gram) signifies a single word and a bigram (2-gram) corresponds to two successive words (denoted as ROUGE-1 and ROUGE-2). ROUGE-L estimates the length of the longest common subsequence (LCS) shared by the model's output and the reference. The ROUGE metric can be formally defined as the following $\text{ROUGE}_\text{Precision}$ and $\text{ROUGE}_\text{Recall}$ under the settings of N-gram and longest common subsequence (LCS):
\begin{equation}
    \begin{cases}
        \notag &\textbf{N-gram Setting:} ~~ \text{ROUGE-N}_\text{Precision} = \frac{|\text{gram}_n(T_\text{ref})~\cap~ \text{gram}_n(T_\text{sum}) | }{| \text{gram}_n(T_\text{ref}) | } , \text{ROUGE-N}_\text{Recall} = \frac{|\text{gram}_n(T_\text{ref})~\cap~ \text{gram}_n(T_\text{sum}) | }{| \text{gram}_n(T_\text{sum}) | } \\\
        \notag &\textbf{LCS Setting:} ~~ \text{ROUGE-L}_\text{Precision} = \frac{|\text{LCS}(T_\text{ref}, T_\text{sum}) | }{| \text{gram}_L(T_\text{ref}) | } , \text{ROUGE-L}_\text{Recall} = \frac{|\text{LCS}(T_\text{ref}, T_\text{sum}) | }{| \text{gram}_L(T_\text{sum}) | }
    \end{cases}
\end{equation}
where $\text{gram}_n(\cdot)$ returns the n-grams set of the input sentence and $\text{LCS}(\cdot,\cdot)$ return the longest common subsequence given the two input sentences, $\text{gram}_L(\cdot)$ returns the L-grams set of the input sentence where L is determined by the length of longest common subsequence in $\text{LCS}(\cdot,\cdot)$. To comprehensively evaluate the performance under the ROUGE metric, we calculate the F1-Score with ROUGE-N$_{\text{F1}} = 2 \times \frac{\text{ROUGE-N}_\text{Precision} \times \text{ROUGE-N}_\text{Recall}}{ \text{ROUGE-N}_\text{Precision} + \text{ROUGE-N}_\text{Recall}}$.

\noindent
\textbf{Implementation Details.} We implement baseline methods and proposed \Algnameabbr{} with Huggingface transformer package~\cite{wolf2019huggingface}. To test the adaptation of \Algnameabbr{} on different distance functions and LLMs, we utilize mean squared error loss and cross-entropy loss with the backbone of \Algnameabbr{}. Specifically, \Algnameabbr{} leverage mean squared error loss with Flan-T5, and utilize cross-entropy loss with BART and Pegasus-xsum. All models are trained until the convergence. 

\begin{table}[t!]
        \centering
        \small
        \setlength{\tabcolsep}{8mm}
        \makebox[1.0\textwidth][c]{
            \resizebox{1.0\textwidth}{!}{
                \begin{tabular}{l | c | c | c }
                    \toprule
                    & ROUGE-1 & ROUGE-2 & ROUGE-L \\
                    \midrule\midrule
                    \multicolumn{4}{c}{\textbf{Pre-trained LLM 1}: Flan-T5~\cite{chung2022scaling}} \\
                    \midrule
                    Base Flan-T5 & 0.5071 & 0.3577 & 0.4798 \\
                    w/ LLM prompts (Mean ~||~ Std.) & 0.5273 ~||~ 0.0081 & 0.3943 ~||~ 0.0080 & 0.4986 ~||~ 0.0079 \\
                    w/ \Algnameabbr{} (Mean ~||~ Std.) & 0.5253 ~||~ 0.0050 & 0.3928 ~||~ 0.0049 & 0.4973 ~||~ 0.0045 \\
                    Std. Deduction (\%) & \textbf{38.2\% $\downarrow$} & \textbf{38.7\% $\downarrow$} &  \textbf{43.1\% $\downarrow$} \\
                    \midrule\midrule
                    \multicolumn{4}{c}{\textbf{Pre-trained LLM 2}: BART~\cite{lewis2020bart}} \\
                    \midrule
                    Base BART & 0.5524 & 0.4244 & 0.5341 \\
                    w/ LLM prompts (Mean ~||~ Std.) & 0.5717 ~||~ 0.0129 & 0.4493 ~||~ 0.0132 & 0.5509 ~||~ 0.0118 \\
                    w/ \Algnameabbr{} (Mean ~||~ Std.) & 0.5636 ~||~ 0.0078 & 0.4341 ~||~ 0.0085 & 0.5429 ~||~ 0.0076\\
                    Std. Deduction. (\%) & \textbf{39.5\% $\downarrow$} & \textbf{35.7\% $\downarrow$} & \textbf{35.6\% $\downarrow$} \\
                    \midrule\midrule
                    \multicolumn{4}{c}{\textbf{Pre-trained LLM 3}: Pegasus-xsum~\cite{zhang2019pegasus}} \\
                    \midrule
                    Base Pegasus-xsum & 0.5503 & 0.4166 & 0.5363 \\
                    w/ LLM prompts (Mean ~||~ Std.) & 0.5677 ~||~ 0.0073 & 0.4441 ~||~ 0.0065 & 0.5493 ~||~ 0.0069 \\
                    w/ \Algnameabbr{} (Mean ~||~ Std.) & 0.5729 ~||~ 0.0053 & 0.4432 ~||~ 0.0058 & 0.5509 ~||~ 0.0056 \\
                    Std. Deduction (\%) & \textbf{27.3\% $\downarrow$} & \textbf{10.8\% $\downarrow$} & \textbf{18.9\% $\downarrow$} \\
                    \bottomrule
                \end{tabular}
            }
        }
        \vspace{-0.2cm}
        \caption{Average F1-score on ROUGE scores and variability of ten LLM-generated prompts.}
        \vspace{0.2cm}
        \label{tab:tb1}
\end{table}

\subsection{Summarization Efficacy and Variability (RQ1)}

\noindent
\textbf{Quantitative Analysis on Summarizing Clinical Notes.} For statistical analysis on summarizing performance, we incorporate proposed \Algnameabbr{} with three different LLMs and evaluate with three different commonly used ROUGE metrics. The F1-score performance results for clinical note summarization tasks are presented through the mean and standard deviation of ten selected LLM-generated prompts under three different LLMs. The performance comparisons are illustrated in Table~\ref{tab:tb1}, where ``w/ LLM prompts" denotes LLMs with ten ChatGPT-generated prompts and ``w/ \Algnameabbr{}" represents our proposed methods. Compared with base LLMs, the proposed \Algnameabbr{} framework outperforms three base LLMs on all metrics. Furthermore, the \Algnameabbr{} framework obtains lower standard deviations than the base LLMs with ChatGPT-generated prompts, indicating the effectiveness of \Algnameabbr{} in mitigating the performance variability. We observe that ChatGPT-generated prompts severely cause performance variability, as the quality of ChatGPT-generated prompts is uncontrollable and predictable. In contrast, \Algnameabbr{} preserves task performance while effectively reducing variability, showing that \Algnameabbr{} avoids an unfavorable trade-off between task performance and variability.

\noindent
\textbf{Case Studies on Clinical Notes Summarization.}
In this section, we conduct case studies to examine the effectiveness of variability using \Algnameabbr{}. The case results are shown in Table~\ref{tab:spec-exmaple} under Flan-T5. In this case study, prompt-1 refers to the prompt ``Please summarize the patient’s medical history and current symptoms"; prompt-2 is ``Can you please provide a brief summary of the patient’s current condition and medical history based on the clinical notes?"; and prompt-3 denotes ``Summarize the following clinical notes." With the same ChatGPT-generated prompts, we observe that \Algnameabbr{} can effectively ensure stable summarizing performance. Particularly, incorrect outcomes (highlighted in red) may arise if \Algnameabbr{} is not utilized, as ChatGPT-generated prompts are not guaranteed to be of sufficient quality for accurately summarizing clinical notes. Prompt-3 notably obtains less information compared to prompt-1 and prompt-2, which causes summarization to miss the points and yield wrong summarization without using proposed \Algnameabbr{}.

\subsection{Performance Variability on Multiple Transformer-based LLMs (RQ2)} 

\begin{figure}[!t]
    \centering
    \subfigure[Flan-T5 Model.]{
    \centering
    \begin{minipage}[t]{0.32\linewidth}
	    \includegraphics[width=0.99\linewidth]{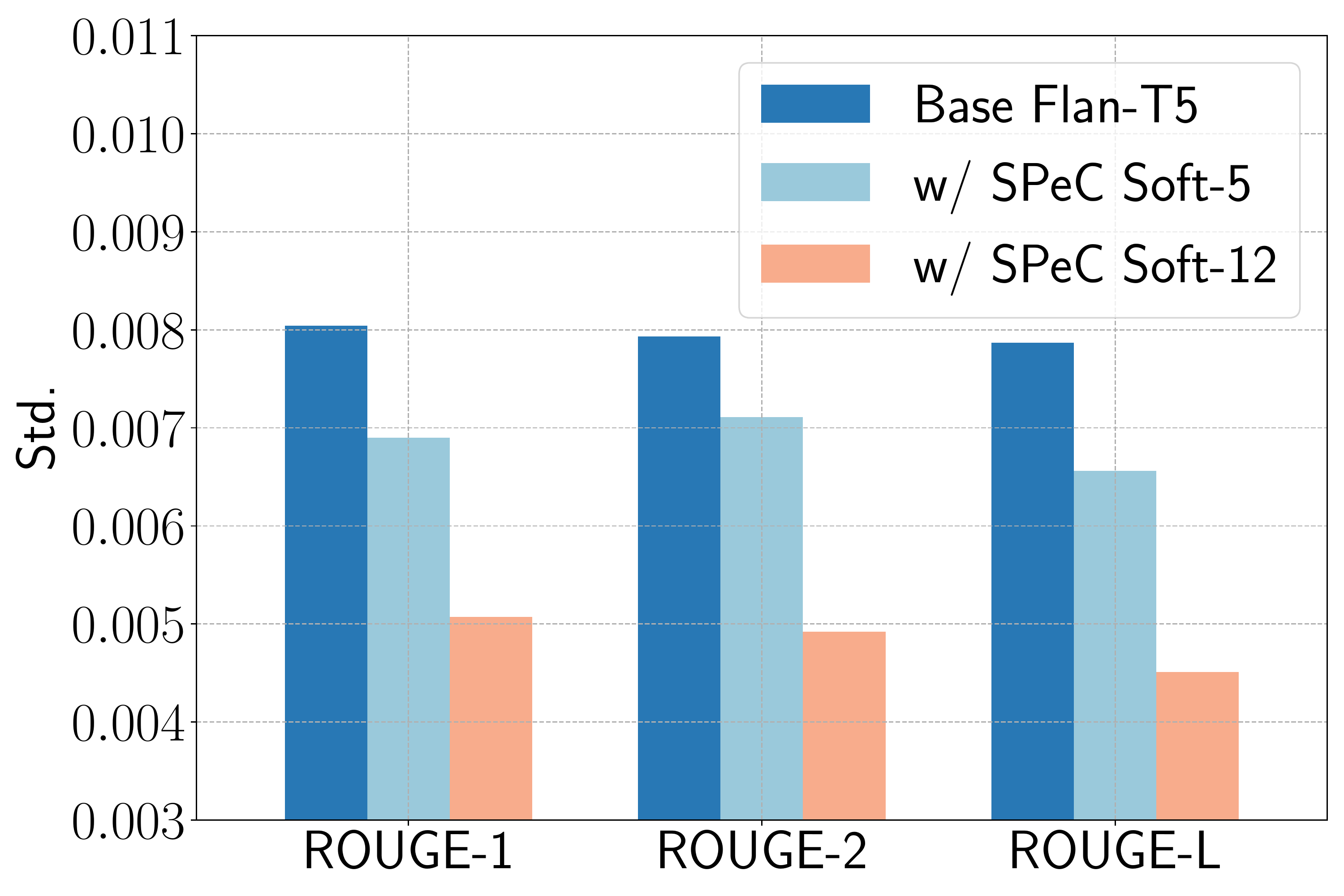}
    \end{minipage}%
    }
    \subfigure[BART Model.]{
    \centering
    \begin{minipage}[t]{0.32\linewidth}
	    \includegraphics[width=0.99\linewidth]{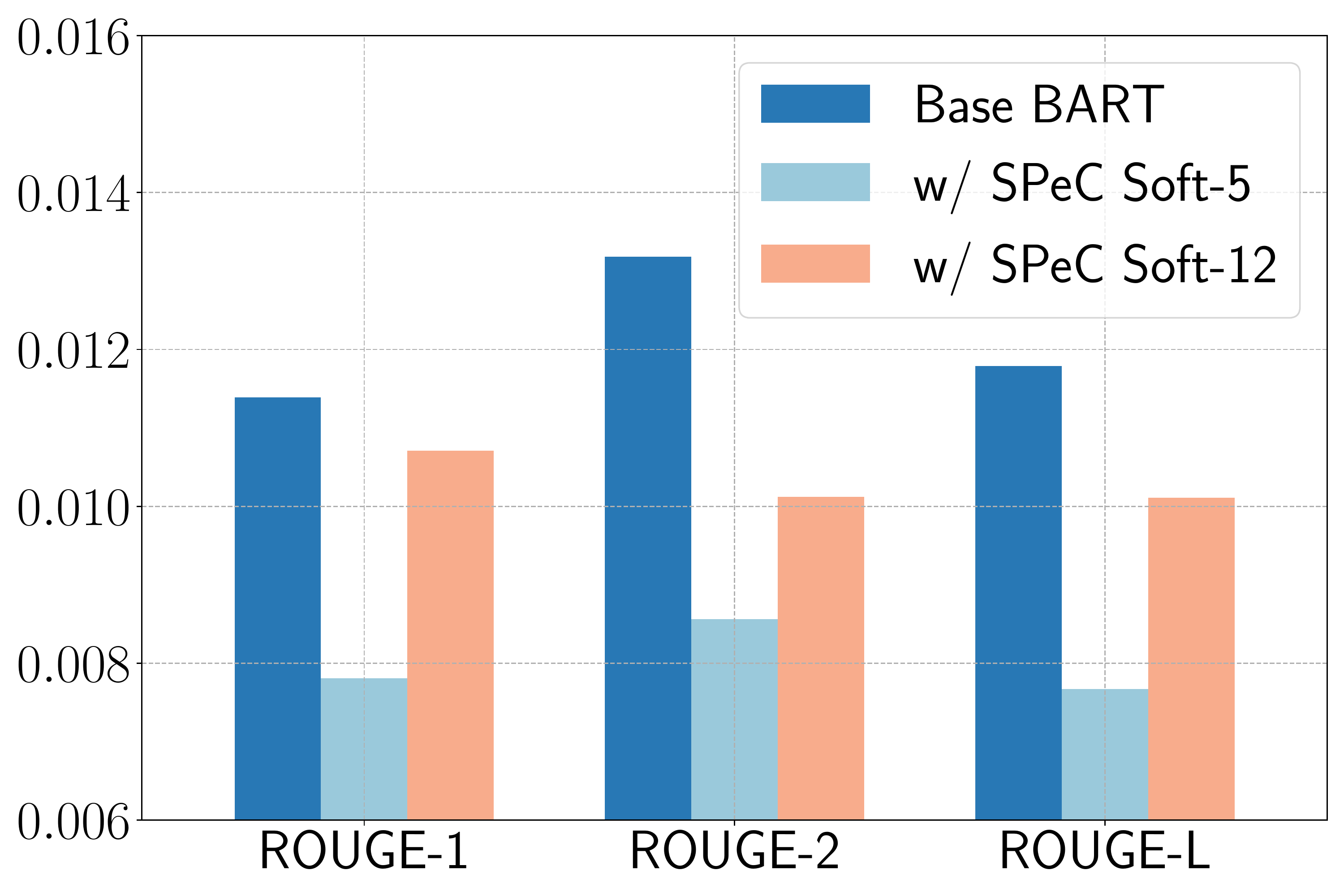}
    \end{minipage}%
    }
    \subfigure[Pegasus-xsum Model.]{
    \centering
    \begin{minipage}[t]{0.32\linewidth}
	    \includegraphics[width=0.99\linewidth]{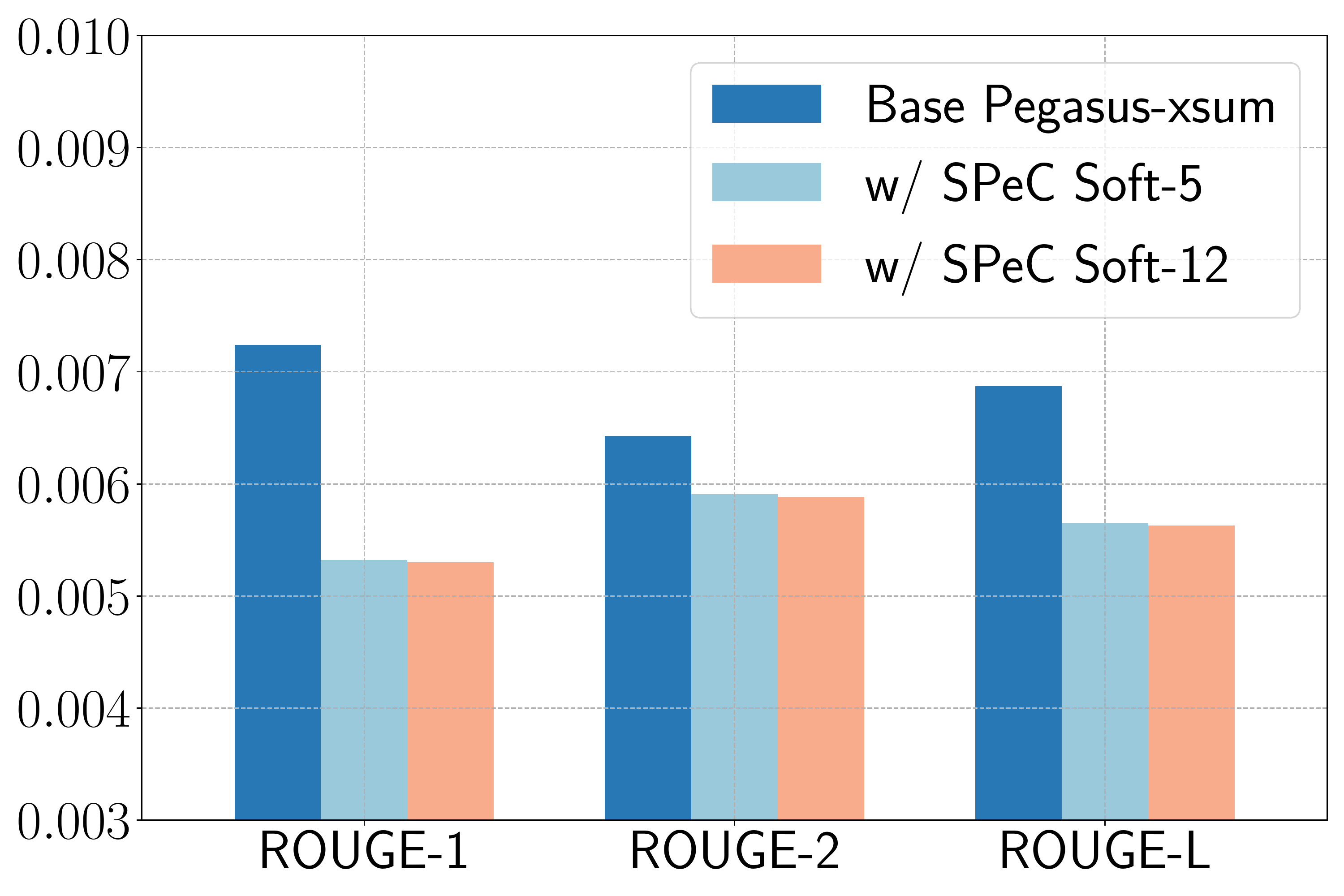}
    \end{minipage}%
    }
    \vspace{-0.4cm}
    \caption{Performance Variability on incorporating three different LLM baselines.}
    \vspace{0.5cm}
    \label{fig:varhist}
\end{figure}

To better understand the effectiveness of \Algnameabbr{} component, we carry out ablation studies to investigate the contributions underlying different sizes of a soft prompt token. In this manner, we can assess the integration capacity of \Algnameabbr{}, which is independent of any potential influence from the length of the soft prompt token. Figure~\ref{fig:varhist} demonstrates the performance variability of \Algnameabbr{}, where "Soft-X" denotes a soft prompt token with a length of X. The ``w/ LLM prompts" denotes the average performances of LLMs with ten different ChatGPT-generated prompts. The principal finding of this section is that the proposed model exhibits lower variability across different incorporated LLMs.

Compared with the standard deviation, we can observe that \Algnameabbr{} framework achieves better performance variability regardless of the soft prompts' length. All three LLMs suffer from the variability brought by ChatGPT-generated prompts. The proposed method can significantly lower the variability up to 43.1\%. This can validate the effectiveness of our model-agnostic \Algnameabbr{}, which can generally mitigate the variability issues caused by LLM-generated prompts.

\begin{table}[t!]
    \centering
    \small
    \begin{tabular}{c | p{13cm}}
        \toprule
        \multicolumn{2}{c}{\textbf{Radiology Report Data}} \\
        \midrule\midrule
        Original Report & Comparing the chest radiograph obtained 4 hours prior, no significant changes are appreciated. Mild pulmonary edema is unchanged. Heart size top-normal. Pleural effusion small, if any. Extensive pleural calcifications somewhat limiting evaluation for pneumonia, though no focal consolidations are identified. \\
        \midrule
        Doctors' Finding & Unchanged, mild pulmonary edema. No radiographic evidence pneumonia.\\
        \midrule
        \multicolumn{2}{c}{\textbf{Clinical Note Summarization w/o \Algnameabbr{}}} \\
        \midrule
        Prompt-1 & Mild pulmonary edema is unchanged \\
        Prompt-2 & \nerbox{Mild congestive heart failure.} \\
        Prompt-3 & Mild pulmonary edema is unchanged. \nerbox{Mild cardiomegaly.} No pneumonia.\\
        \midrule
        \multicolumn{2}{c}{\textbf{Clinical Note Summarization w/ \Algnameabbr{}}} \\
        \midrule
        Prompt-1 & Mild pulmonary edema, unchanged. No focal consolidation. Extensive pleural calcifications. \\
        Prompt-2 & Mild pulmonary edema, unchanged. Small, if any, pleural effusion. Extensive pleural calcifications limits evaluation for pneumonia. \\
        Prompt-3 & Mild pulmonary edema, slightly improved compared to the prior exam. \\
        \bottomrule
    \end{tabular}
    \caption{Performance variability comparison of Flan-T5 w/ utilizing \Algnameabbr{} and w/o exploiting \Algnameabbr{}.}
    \vspace{0.3cm}
    \label{tab:spec-exmaple}
\end{table}

\subsection{Ablation Studies on Soft Prompt Token (RQ3)} 

In order to further verify the effectiveness of \Algnameabbr{} when using different soft prompt tokens, we conducted ablation studies to investigate the contribution of the soft prompt token selection. The ablation studies are conducted to compare the performance of selecting an in-distribution soft prompt token and an out-of-distribution soft prompt token. In the previous experiment sections, we adopt an in-distribution soft prompt token, which is \textit{``radiologist describe stable normality and abnormality exam"}, in \Algnameabbr{}. Some research has shown that Bayesian optimization can effectively deduce the uncertainty and variability of the model predictions~\cite{gal2016dropout, lakshminarayanan2017simple}. The selection of in-distribution tokens can be seen as a Bayesian optimization process that finds the maximum likelihood estimation of a given soft token encoder, which can significantly reduce the variability and prevent the soft token from poisoning the prior embedding distribution.
Unlike the in-distribution soft prompt tokens, out-of-distribution soft prompt tokens, such as ``\#\#1" and ``\#\#2", are not in the token dictionary of the given LLMs. 

Table~\ref{tab:tb2} demonstrates the comparison between two types of soft prompt token selection under the Flan-T5 model. We observe that the in-distribution tokens achieve better performance than out-of-distribution ones, successfully deducing the performance variability and maintaining the summarizing performance on the clinical notes dataset. This phenomenon consistently occurs under three different ROUGE metrics, where in-distribution ones offer better variability deduction up to 43.1\% compared to out-of-distribution ones with up to 22.5\%. In this manner, we can provide the selection reference, where the soft prompt tokens are recommended to be the in-distribution tokens instead of generating the randomly-created words.

\begin{table}[t!]
        \centering
        \small
        \setlength{\tabcolsep}{7mm}
        \makebox[1.0\textwidth][c]{
            \resizebox{0.95\textwidth}{!}{
                \begin{tabular}{l | c | c | c }
                    \toprule
                    & ROUGE-1 & ROUGE-2 & ROUGE-L \\
                    \midrule\midrule
                    w/ LLM prompts (Mean) & 0.5273 & 0.3943 & 0.4986 \\
                    w/ LLM prompts (Std.) & 0.0081 & 0.0080 & 0.0079 \\
                    \midrule\midrule
                    \multicolumn{4}{c}{\textbf{Case 1:} In-distribution soft prompt token} \\
                    \midrule
                    w/ \Algnameabbr{} (Mean) & 0.5253 & 0.3928 & 0.4973 \\
                    w/ \Algnameabbr{} (Std.)  & 0.0050 & 0.0049 & 0.0045 \\
                    Deduction (Mean\% ~||~ Std.\%) & \textbf{0.3\% $\downarrow$} ~||~  \textbf{38.2\% $\downarrow$} &  \textbf{0.3\% $\downarrow$} ~||~  \textbf{38.7\% $\downarrow$} &  \textbf{0.2\% $\downarrow$} ~||~  \textbf{43.1\% $\downarrow$} \\
                    \midrule\midrule
                    \multicolumn{4}{c}{\textbf{Case 2:} Out-of-distribution soft prompt token} \\
                    \midrule
                    w/ \Algnameabbr{} (Mean) & 0.5185 & 0.3853 & 0.4890 \\
                    w/ \Algnameabbr{} (Std.) & 0.0065 & 0.0062 & 0.0066 \\
                    Deduction (Mean\% ~||~ Std.\%) & 1.6\% $\downarrow$ ~||~ 19.7\% $\downarrow$ &   2.3\% $\downarrow$ ~||~ 22.5\% $\downarrow$ & 2.3\% $\downarrow$ ~||~ 16.4\% $\downarrow$\\
                    \bottomrule
                \end{tabular}
            }
        }
        \vspace{-0.2cm}
        \caption{Performance Comparison between in-distribution soft prompt token and out-of-distribution soft prompt token.}
        \label{tab:tb2}
\end{table}

\section{Limitations and Ethics Statements}
In this work, we verify our SPeC framework on the radiology reports for initial testing. While using the SPeC model to summarize radiology reports can be beneficial, several limitations should be considered to be further adaptive to other kinds of medical reports. For instance, more complex medical terminologies and abbreviations might appear in other medical reports, which can lead to vague or misleading summaries when the model attempts to generate concise information regarding non-radiology reports. Some medical reports may reveal more private and demographic information about the patients for necessity. Ensuring patient privacy and complying with data security regulations may be more challenging than the scenario on radiology report summarization. 

The ethics statements are as follows. Despite the significant variance deduction from the SPeC framework, relying solely on AI-generated summaries might lead to reduced human oversight, potentially overlooking critical findings or errors that only a human expert could detect. Real-world utilization is highly encouraged to be double-verified by well-trained doctors or physicians. Furthermore, we are aware that MIMIC-CXR is not allowed to directly send to the OpenAI-related platforms. In this work, we did not input any reports from the MIMIC-CXR dataset on the OpenAI-related platforms, such as ChatGPT and GPT4, to generate any results. Instead, we only input the unrelated question to OpenAI-related platforms and get the LLM-generated prompts for conducting our experiments. The access right of publicly available MIMIC-CXR datasets is authorized by the data providers, and the researchers in this work have completed the required training from CITI training courses.
Finally, using medical datasets for AI research requires strict adherence to ethical principles. Researchers in this work are highly concerned about prioritizing patient privacy, informed consent, and data fairness while ensuring responsible usage of AI models. The ultimate goal should be to benefit patients and society by improving healthcare outcomes responsibly and collaboratively.

\section{Conclusion}
This study introduces a new model-agnostic pipeline, \Algnameabbr{} (\Algnameunderline{}), designed to address the issue of performance variability in clinical note summarization tasks, particularly when leveraging large language models. By employing soft prompts in conjunction with discrete prompts, our approach effectively mitigates the variance in summarizing clinical notes while still harnessing the benefits of prompt-based summarization techniques. The success of \Algnameabbr{} framework in delivering consistent and dependable summaries of medical information is a significant contribution to the field, empowering healthcare practitioners to make informed decisions and provide optimal patient care. The study primarily focuses on radiology report summarization, as these reports adhere to a highly standardized format. Moving forward, we plan to extend the application of this method to encompass more complex clinical note summarization tasks.


\makeatletter
\renewcommand{\@biblabel}[1]{\hfill #1.}
\makeatother

\bibliographystyle{vancouver}
\bibliography{amia}

\begin{thebibliography}{10}

\bibitem{wagner2020augmented}
Wagner T, Shweta F, Murugadoss K, Awasthi S, Venkatakrishnan A, Bade S, et~al.
\newblock Augmented curation of clinical notes from a massive EHR system
  reveals symptoms of impending COVID-19 diagnosis.
\newblock Elife. 2020;9:e58227.

\bibitem{pivovarov2015automated}
Pivovarov R, Elhadad N.
\newblock Automated methods for the summarization of electronic health records.
\newblock Journal of the American Medical Informatics Association.
  2015;22(5):938-47.

\bibitem{wang2021systematic}
Wang M, Wang M, Yu F, Yang Y, Walker J, Mostafa J.
\newblock A systematic review of automatic text summarization for biomedical
  literature and EHRs.
\newblock Journal of the American Medical Informatics Association.
  2021;28(10):2287-97.

\bibitem{gershanik2011critical}
Gershanik EF, Lacson R, Khorasani R.
\newblock Critical finding capture in the impression section of radiology
  reports.
\newblock In: AMIA Annual Symposium Proceedings. vol. 2011. American Medical
  Informatics Association; 2011. p. 465.

\bibitem{cai2021chestxraybert}
Cai X, Liu S, Han J, Yang L, Liu Z, Liu T.
\newblock Chestxraybert: A pretrained language model for chest radiology report
  summarization.
\newblock IEEE Transactions on Multimedia. 2021.

\bibitem{xiao2018opportunities}
Xiao C, Choi E, Sun J.
\newblock Opportunities and challenges in developing deep learning models using
  electronic health records data: a systematic review.
\newblock Journal of the American Medical Informatics Association.
  2018;25(10):1419-28.

\bibitem{choi2018mime}
Choi E, Xiao C, Stewart W, Sun J.
\newblock Mime: Multilevel medical embedding of electronic health records for
  predictive healthcare.
\newblock Advances in neural information processing systems. 2018;31.

\bibitem{wei2022emergent}
Wei J, Tay Y, Bommasani R, Raffel C, Zoph B, Borgeaud S, et~al.
\newblock Emergent abilities of large language models.
\newblock arXiv preprint arXiv:220607682. 2022.

\bibitem{liu2023pre}
Liu P, Yuan W, Fu J, Jiang Z, Hayashi H, Neubig G.
\newblock Pre-train, prompt, and predict: A systematic survey of prompting
  methods in natural language processing.
\newblock ACM Computing Surveys. 2023;55(9):1-35.

\bibitem{zhang2022neural}
Zhang Y, Zhou K, Liu Z.
\newblock Neural prompt search.
\newblock arXiv preprint arXiv:220604673. 2022.

\bibitem{raffel2020exploring}
Raffel C, Shazeer N, Roberts A, Lee K, Narang S, Matena M, et~al.
\newblock Exploring the Limits of Transfer Learning with a Unified Text-to-Text
  Transformer.
\newblock Journal of Machine Learning Research. 2020;21:1-67.

\bibitem{lester2021power}
Lester B, Al-Rfou R, Constant N.
\newblock The power of scale for parameter-efficient prompt tuning.
\newblock arXiv preprint arXiv:210408691. 2021.

\bibitem{chung2022scaling}
Chung HW, Hou L, Longpre S, Zoph B, Tay Y, Fedus W, et~al.
\newblock Scaling instruction-finetuned language models.
\newblock arXiv preprint arXiv:221011416. 2022.

\bibitem{gupta2010survey}
Gupta V, Lehal GS.
\newblock A survey of text summarization extractive techniques.
\newblock Journal of emerging technologies in web intelligence.
  2010;2(3):258-68.

\bibitem{lin2019abstractive}
Lin H, Ng V.
\newblock Abstractive summarization: A survey of the state of the art.
\newblock In: Proceedings of the AAAI Conference on Artificial Intelligence.
  vol.~33; 2019. p. 9815-22.

\bibitem{see2017get}
See A, Liu PJ, Manning CD.
\newblock Get To The Point: Summarization with Pointer-Generator Networks.
\newblock In: Proceedings of the 55th Annual Meeting of the Association for
  Computational Linguistics (Volume 1: Long Papers); 2017. p. 1073-83.

\bibitem{lewis2020bart}
Lewis M, Liu Y, Goyal N, Ghazvininejad M, Mohamed A, Levy O, et~al.
\newblock BART: Denoising Sequence-to-Sequence Pre-training for Natural
  Language Generation, Translation, and Comprehension.
\newblock In: Proceedings of the 58th Annual Meeting of the Association for
  Computational Linguistics; 2020. p. 7871-80.

\bibitem{jain2022survey}
Jain R, Jangra A, Saha S, Jatowt A.
\newblock A Survey on Medical Document Summarization.
\newblock arXiv preprint arXiv:221201669. 2022.

\bibitem{Zhou2022-td}
Zhou Y, Muresanu AI, Han Z, Paster K, Pitis S, Chan H, et~al.
\newblock Large language models are human-level prompt engineers.
\newblock arXiv preprint arXiv:221101910. 2022.

\bibitem{shin2020autoprompt}
Shin T, Razeghi Y, Logan~IV RL, Wallace E, Singh S.
\newblock Autoprompt: Eliciting knowledge from language models with
  automatically generated prompts.
\newblock arXiv preprint arXiv:201015980. 2020.

\bibitem{li2021prefix}
Li XL, Liang P.
\newblock Prefix-tuning: Optimizing continuous prompts for generation.
\newblock arXiv preprint arXiv:210100190. 2021.

\bibitem{wang2022promptehr}
Wang Z, Sun J.
\newblock PromptEHR: Conditional Electronic Healthcare Records Generation with
  Prompt Learning.
\newblock arXiv preprint arXiv:221101761. 2022.

\bibitem{vu2021spot}
Vu T, Lester B, Constant N, Al-Rfou R, Cer D.
\newblock Spot: Better frozen model adaptation through soft prompt transfer.
\newblock arXiv preprint arXiv:211007904. 2021.

\bibitem{kim2023prompt}
Kim M, Kim HI, Ro YM.
\newblock Prompt Tuning of Deep Neural Networks for Speaker-adaptive Visual
  Speech Recognition.
\newblock arXiv preprint arXiv:230208102. 2023.

\bibitem{zhang2023pheme}
Zhang S, Li H, Tang R, Ding S, Rasmy L, Zhi D, et~al.
\newblock PheME: A deep ensemble framework for improving phenotype prediction
  from multi-modal data.
\newblock arXiv preprint arXiv:230310794. 2023.

\bibitem{yuan2023llm}
Yuan J, Tang R, Jiang X, Hu X.
\newblock LLM for Patient-Trial Matching: Privacy-Aware Data Augmentation
  Towards Better Performance and Generalizability.
\newblock arXiv preprint arXiv:230316756. 2023.

\bibitem{Arora2022-dj}
Arora S, Narayan A, Chen MF, Orr LJ, Guha N, Bhatia K, et~al.
\newblock Ask Me Anything: A simple strategy for prompting language models.
\newblock arXiv preprint arXiv:221002441. 2022.

\bibitem{Wu2022-dw}
Wu T, Jiang E, Donsbach A, Gray J, Molina A, Terry M, et~al.
\newblock Promptchainer: Chaining large language model prompts through visual
  programming.
\newblock In: CHI Conference on Human Factors in Computing Systems Extended
  Abstracts; 2022. p. 1-10.

\bibitem{Mishra2021-bl}
Mishra S, Khashabi D, Baral C, Choi Y, Hajishirzi H.
\newblock Reframing Instructional Prompts to GPTk's Language.
\newblock arXiv preprint arXiv:210907830. 2021.

\bibitem{johnson2019mimic}
Johnson A, Lungren M, Peng Y, Lu Z, Mark R, Berkowitz S, et~al.
\newblock MIMIC-CXR-JPG-chest radiographs with structured labels (version 2.0.
  0).
\newblock PhysioNet. 2019;10:8360-t248.

\bibitem{kahn2009toward}
Kahn~Jr CE, Langlotz CP, Burnside ES, Carrino JA, Channin DS, Hovsepian DM,
  et~al.
\newblock Toward best practices in radiology reporting.
\newblock Radiology. 2009;252(3):852-6.

\bibitem{gpt4}
OpenAI.
\newblock GPT-4 Technical Report.
\newblock arXiv preprint arXiv:230308774. 2023.

\bibitem{zhang2019pegasus}
Zhang J, Zhao Y, Saleh M, Liu PJ.
\newblock PEGASUS: Pre-training with Extracted Gap-sentences for Abstractive
  Summarization.
\newblock arXiv preprint arXiv:191208777. 2019.

\bibitem{lin2004rouge}
Lin CY.
\newblock Rouge: A package for automatic evaluation of summaries.
\newblock In: Text summarization branches out; 2004. p. 74-81.

\bibitem{wolf2019huggingface}
Wolf T, Debut L, Sanh V, Chaumond J, Delangue C, Moi A, et~al.
\newblock Huggingface's transformers: State-of-the-art natural language
  processing.
\newblock arXiv preprint arXiv:191003771. 2019.

\bibitem{gal2016dropout}
Gal Y, Ghahramani Z.
\newblock Dropout as a bayesian approximation: Representing model uncertainty
  in deep learning.
\newblock In: international conference on machine learning. PMLR; 2016. p.
  1050-9.

\bibitem{lakshminarayanan2017simple}
Lakshminarayanan B, Pritzel A, Blundell C.
\newblock Simple and scalable predictive uncertainty estimation using deep
  ensembles.
\newblock Advances in neural information processing systems. 2017;30.

\end{thebibliography}

\end{document}